\documentclass[a4paper,conference]{IEEEtran}
\usepackage{times}
\usepackage{epsfig}
\usepackage{graphicx}
\usepackage{amsmath}
\usepackage{amssymb}

\usepackage{algorithm,algorithmic}
\usepackage{float}
\usepackage{subcaption} 
\usepackage{tabularx}

\def\arrvline{\hfil\kern\arraycolsep\vline\kern-\arraycolsep\hfilneg}

\ifCLASSINFOpdf
\else
\fi
\begin{document}
%
\title{Local Facial Attribute Transfer through Inpainting}

\author{Ricard Durall$^{1,2,3}$ \qquad Franz-Josef Pfreundt$^{1}$ \qquad
Janis Keuper$^{1,4}$\\
$^1$Fraunhofer ITWM, Germany\\
$^2$IWR, University of Heidelberg, Germany\\
$^3$Fraunhofer Center Machine Learning, Germany\\
$^4$Institute for Machine Learning and Analytics, Offenburg University, Germany\\
}

%


\maketitle

\begin{abstract}
The term ``attribute transfer'' refers to the tasks of altering images in such a way, that the semantic interpretation of a given input image is 
shifted towards an intended direction, which is quantified by semantic attributes. Prominent example applications are photo realistic changes of facial 
features and expressions, like changing the hair color, adding a smile, enlarging the nose or altering the entire context of a scene, like transforming a 
summer landscape into a winter panorama.
Recent advances in attribute transfer are mostly based on generative deep neural networks, using various techniques to manipulate images in the latent space of the 
generator.\\
In this paper, we present a novel method for the common sub-task of local attribute transfers, where only parts of a face have to be altered in order to achieve semantic
changes (e.g. removing a \textit{mustache}). In contrast to previous methods, where such local changes have been implemented by generating new (global) images, we propose to formulate local attribute
transfers as an inpainting problem. Removing and regenerating only parts of images, our ``Attribute  Transfer Inpainting Generative Adversarial Network'' (ATI-GAN) is able to utilize local context information to focus on the attributes while keeping the background unmodified resulting in visually sound results.
\end{abstract}


%

\section{Introduction}
Generative deep learning modeling is an ongoing growing field, in which recent works have shown remarkable
success in different domains. In particular, the computer vision community has witnessed a dramatic improvement a in large variety of tasks, ranging from image synthesis 
\cite{zhang2017stackgan,karras2017progressive,brock2018large} to
image-to-image translation
\cite{iizuka2017globally,choi2018stargan,mo2018instanceaware}. The
latter task poses the problem of translating images from one domain to
another, including
style transfer \cite{isola2017image,zhu2017unpaired,karras2019style}, inpainting
\cite{yeh2017semantic,li2017generative,yu2018generative,yu2019free},
attribute transfer \cite{kim2017learning,creswell2017adversarial,choi2018stargan}, and others. 

\begin{figure}[t]
\centering
   \includegraphics[width=\linewidth]{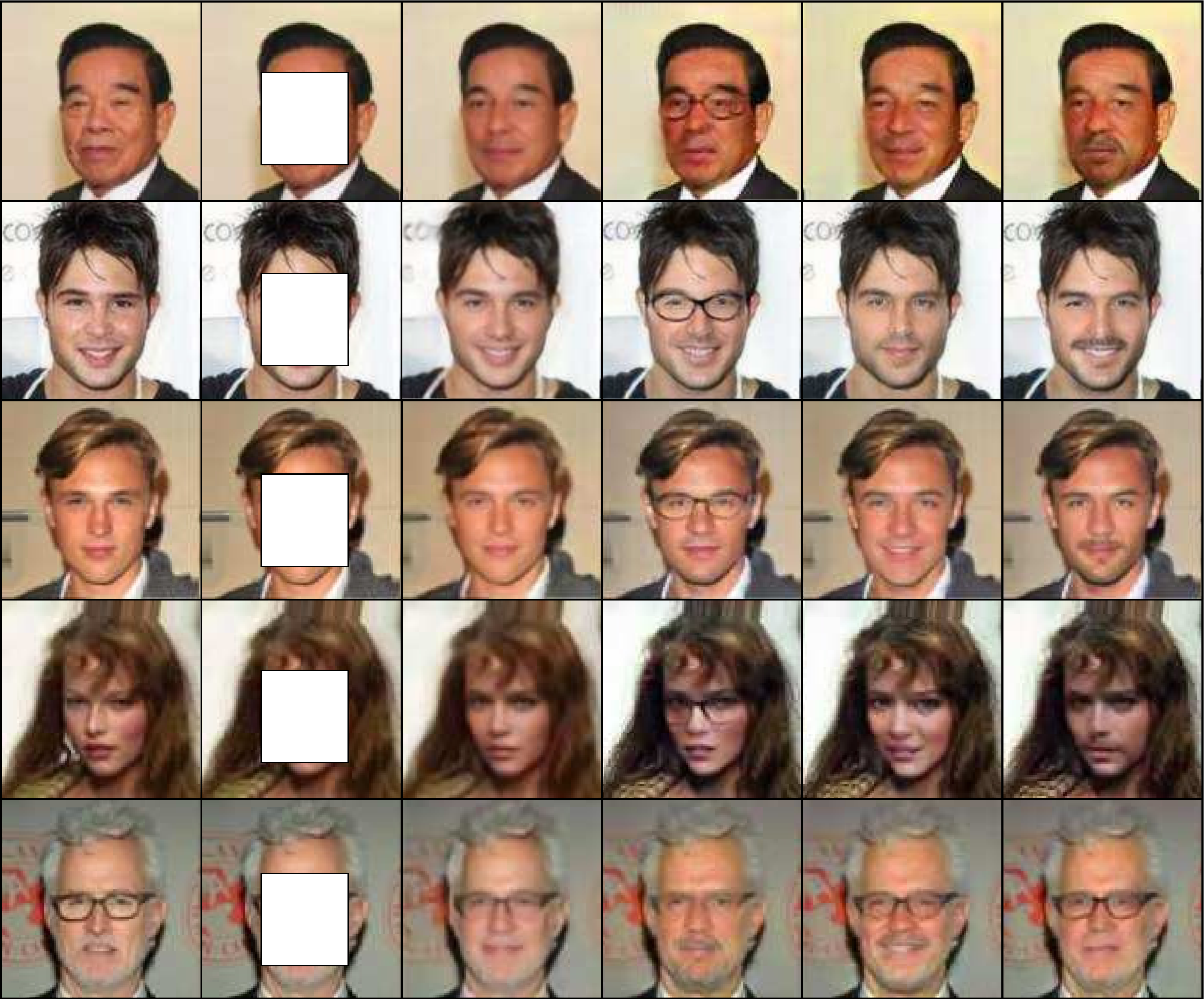}
   \caption{Image-to-image translation results on the CelebA dataset. The first column shows the original images, the second the input masked images, the third the inpainted translation and the remaining columns are attribute transfer results (\textit{eyeglasses}, \textit{smiling} and \textit{mustache}). Note that results from attribute translation are opposite of the original. }
\label{fig:main}
\end{figure}

The objective of attribute transfer is to synthesize new and realistic appearing images for a
pre-defined target domain. For instance, Fig. \ref{fig:main} column 1 row 5 shows a
\textit{non-smiling} man with \textit{mustache} wearing \textit{eyeglasses} (these are the given
attributes, a.k.a. domains) and the output results show how these attributes have been changed one at a time according to our attribute target domain. We refer to a domain as a set of images sharing the same
attributes. Such attributes are meaningful semantic features of an image, such as a \textit{mustache} or a face with \textit{eyeglasses}. Like other image-to-image translations, attribute transfer
methods have also achieved an impressive progress by implementing different variants of GANs
\cite{creswell2017adversarial,choi2018stargan,karras2019style} leading to state-of-the-art
results in the field. Nevertheless, these attribute transfer approaches are mostly based on  the
global manipulation of GAN latent space. As a result, in order to produce good transfer results, the
aforementioned methods require additional inverse generator paths (which tends to make them less
stable) and can be quite cumbersome.

Image inpainting or completion refers to the task of inferring locally missing or damaged parts of an image. It has been applied to many different applications like photo editing, restoration of damaged
paintings, image-based rendering and computational photography.
The main challenge of image inpainting is to synthesize realistic pixels for the missing regions
that are coherent with existing ones. Image inpainting techniques are mostly separated into two
groups regarding their basic approach. The first group uses local methods  \cite
{barnes2009patchmatch,hu2013fast} based on low-level feature information, such
as color or texture, to attempt to solve the problem. The second group relies on recognizing
patterns in images, e.g. deep convolutional neural networks (CNN), to predict pixels for the missing
regions. CNN-based models deal with both local and global features and can, in combination with
generative adversarial networks (GANs) \cite{goodfellow2014generative}, produce realistic inpainted
outputs. The introduction of GANs have inspired recent works
\cite{pathak2016context,yeh2017semantic,li2017generative,iizuka2017globally,yu2018generative} which
formulate the inpainting tasks as a conditional image generation problem, using a generator for
inpainting  and a discriminator for evaluating the result.\\


\noindent \textbf{Contributions.}
The contribution of this paper is a novel attribute transfer approach that alters given natural images in such a way, that the output image meets the pre-defined visual attributes. To do so, our proposed architecture integrates an inpainting block. This block allows to focus all the attention into the region of interest (where the attributes are located) while keeping the background unmodified but still consistent with the changes.
In particular, we take advantage of the fact that most facial attributes are induced by local structures (e.g. relative position between eyes and ears). Hence, it is sufficient to change only a few parts of the face and to force the generator to integrate them into realistic outputs. Note that the hole/mask (in terms of inpainting) is generated by our method to apply this ``trick''.
The proposed ATI-GAN model integrates inpainting for local attribute transfer in a single end-to-end architecture with three main building blocks: on one hand, we have an inpainting network
that takes masked images as input and outputs realistically restored images; on the other hand, we have
a second network that takes these inpainted images and encoded attributes (e.g. one-hot vector) as
input, and learns how to separate attribute information from the rest of the image representation; finally, a third network, which acts as a discriminator, judges if the overall result looks realistic.

Evaluations of our model on the CelebA \cite{liu2015deep} dataset of faces demonstrate the capacity of ATI-GAN to produce high quality outputs. Quantitative and qualitative results show superior inpainting and attribute transfer performance.





\section{Related Work}

In computer vision. deep learning approaches have heavily contributed in many semantic
image understanding tasks. In this section, we briefly review publications
related to our work in each of the different sub-fields. In particular, since
our proposal is based on GANs and image translation, more specifically in
inpainting and attribute transfer, we will review the seminal work in that
direction. 

\subsection{Generative Adversarial Networks} 
Generative adversarial networks \cite{goodfellow2014generative} have arisen as
a reliable framework for deep generative models. They have shown remarkable
results in various computer vision tasks like image generation
\cite{zhang2017stackgan,karras2017progressive}, style transfer
\cite{isola2017image,zhu2017unpaired},
inpainting
\cite{pathak2016context,yeh2017semantic,li2017generative,iizuka2017globally,yu2018generative}
and attribute transfer
\cite{kim2017learning,creswell2017adversarial,choi2018stargan}. The vanilla GAN
model consists of two networks, a generator $G$ and a discriminator $D$. Its
procedure can be seen as a minmax game between $G$, which learns how to
generate samples resembling real data, and a discriminator $D$, which learns to
discriminate between real and fake data. Throughout this process, $G$
indirectly learns how to model the input image distribution
$p_{\mathrm{data}}$ by taking samples $\mathrm{\mathbf{z}}$ from a fixed
distribution $p_{\mathrm{z}}$ (e.g. Gaussian) and forcing the generated samples
$G(\mathrm{\mathbf{z}})$ to match the input images $\mathrm{\mathbf{x}}$. The
objective loss function is defined as
\begin{align}
\begin{split}
	\min_{G} \max_{D} \mathcal{L}(D,G) =\,& \mathbb{E}_{\mathrm{\mathbf{x}} \sim p_{\mathrm{data}}} \left[ \log \left(D(\mathrm{\mathbf{x}})\right) \right] \,+\\	
	+\,& \mathbb{E}_{\mathrm{\mathbf{z}} \sim p_{\mathrm{z}}}[\log(1-D(G(\mathrm{\mathbf{z}})))].
\end{split}
\end{align}

\vspace{3mm}GAN-based conditional approach \cite{mirza2014conditional} has shown a rapid progress and it has become an essential ingredient for recent research. The intuition behind this kind of GANs is to insert  the class information into the model, in order to generate samples that are conditioned on the class. In this work, we take advantage of this property and we encode the attribute characteristics as a conditional information which will be fed into the model.

\begin{figure*}
\begin{subfigure}{\linewidth}
\centering
\includegraphics[width=.60\linewidth]{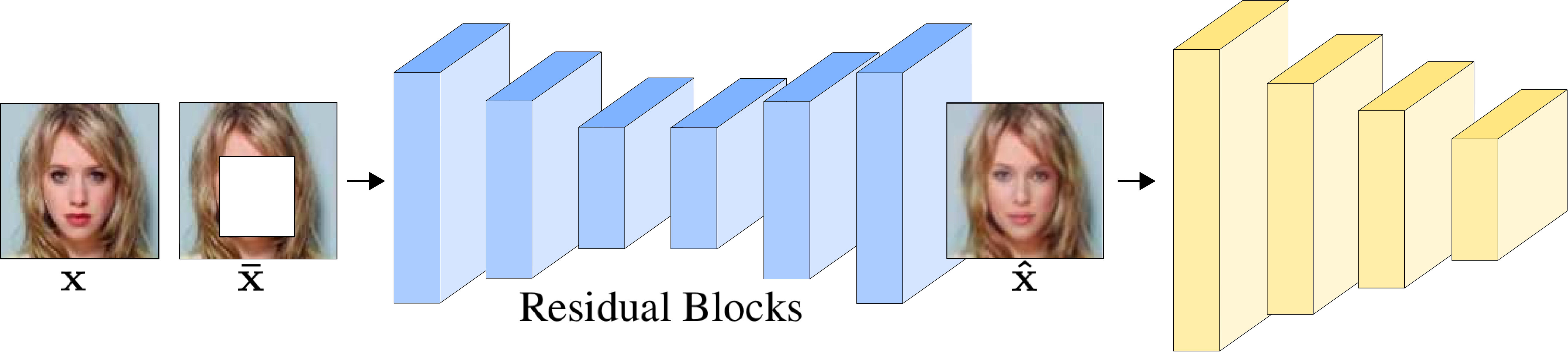}
\caption{Training the reconstructor $R$.}
\label{fig:r}
\end{subfigure}\\[2ex]
\begin{subfigure}{\linewidth}
\centering
\includegraphics[width=.68\linewidth]{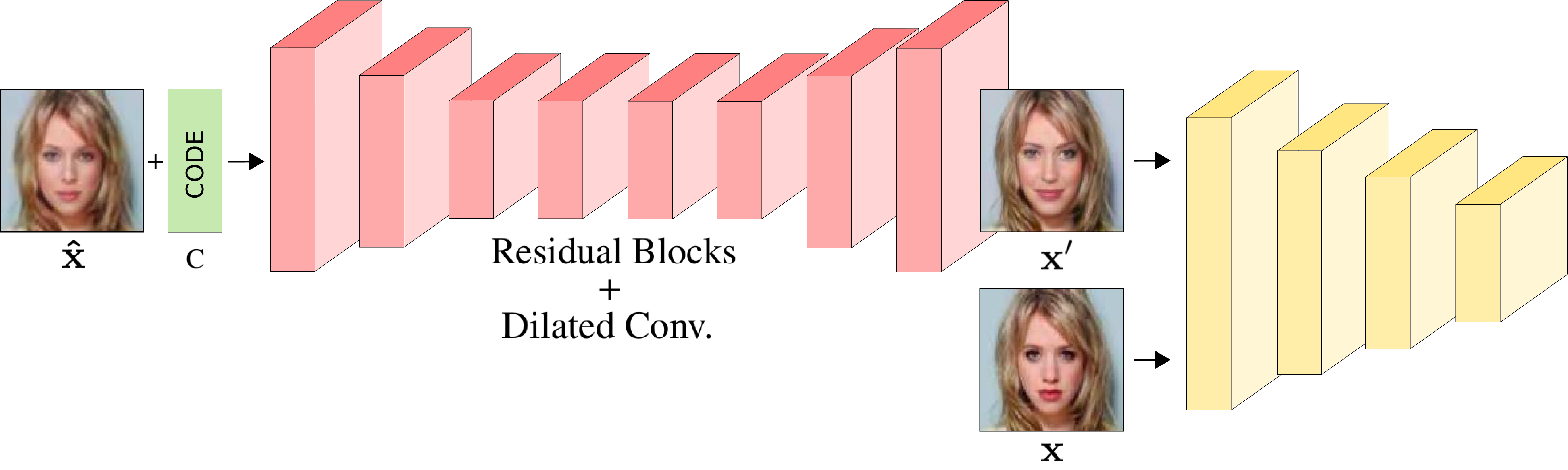}
\caption{Training the discriminator $D$.}
\label{fig:d}
\end{subfigure}\\[2ex]
\begin{subfigure}{\linewidth}
\centering
\includegraphics[width=1\linewidth]{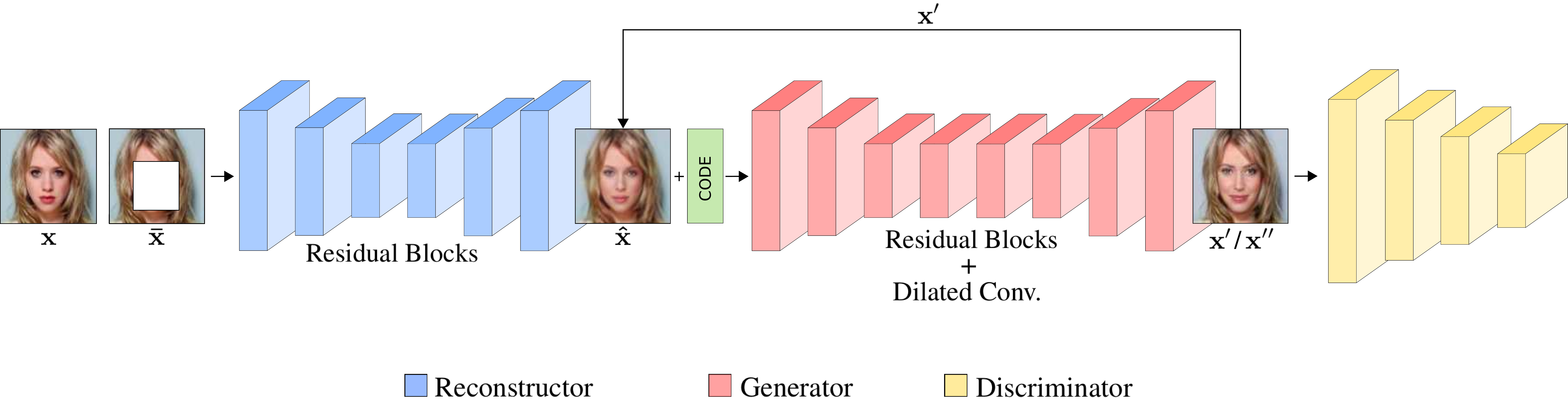}
\caption{Training the generator $G$.}
\label{fig:g}
\end{subfigure}
\caption{Overview of ATI-GAN network while training. It is composed of a reconstructor $R$, a generator $G$ and a discriminator $D$ which are trained independently. (\subref{fig:r}) $R$ takes the masked image as input and reconstructs a realistic image, which is judged by $D$. (\subref{fig:d}) $D$ takes real and fake images as inputs and learns to distinguish between them at global and patch levels. Furthermore, it also learns to classify them to their corresponding domain. (\subref{fig:g}) $G$ takes both, the output from $R$ and the target domain label, as  inputs and generates the domain transformed image. Then, the transformed image and the original domain label are fed into $G$ (creating a loop). After that, both outputs from $G$ (from the two iterations) are passed to $D$ one at a time.}
\label{fig:all}
\end{figure*}

\subsection{Image Inpainting} 
Classical inpainting methods are often based on either local or non-local
information to rebuild the patches. Local methods \cite
{barnes2009patchmatch,hu2013fast} attempt to solve the
problem using only context information, such as color or texture, i.e. matching, copying and merging backgrounds patches into holes by propagating the information from
hole boundaries. These approaches need very little training or prior knowledge,
and provide good results especially in background inpainting tasks. However,
they do not perform well for large patches because of their inability of
generating novel image contents. More powerful methods are global,
content-based and semantic inpainting approaches. Even though, these
techniques require more expensive dense patch computations, they can handle larger patches successfully. 
In many models, CNN-based approaches have become the de
facto implementation due to their capability to learn to recognize patterns in
images and use them to fill holes in images.

GANs for image translation
\cite{pathak2016context,yeh2017semantic,li2017generative,iizuka2017globally}
have emerged as a promising paradigm for inpainting tasks. Nowadays, they are already able to produce realistic synthetic outputs with high quality image resolution \cite{yu2018generative,yu2019free}. In order to reach this point though, GAN's techniques have been evolving quite intensively over the past few years. Initially, the inpainting task was formulated as a conditional image generation problem, consisting of one generator and one
discriminator. However, more recent works \cite{iizuka2017globally,yu2018generative} have introduced the
concept of global and local discriminators. Furthermore, apart from modifying the topology from the discriminator, recent methods  \cite{iizuka2017globally,li2017generative,pathak2016context,yu2018generative} have adopted new losses such as Wasserstein \cite{arjovsky2017wasserstein} or Wasserstein with gradient penalty  \cite{gulrajani2017improved}. Inspired by all of these works, our approach also leverages the conditioned
adversarial framework with global and local discriminators together with Wasserstein loss gradient penalty.

\subsection{Attribute Transfer}
Small and unbalanced datasets can cause severe problems when training a machine learning model.
Recently, numerous works have put their attention towards transferring visual
attributes, such as color \cite{zhang2016colorful}, texture
\cite{isola2017image,zhu2017unpaired}, facial
features
\cite{kim2017learning,creswell2017adversarial,choi2018stargan}
and more, for data augmentation. However, although most of the approaches correctly synthesize new attributes belonging to the target
domain, it is still very challenging to generalize attributes between different applications since they are usually designed to transfer a specific type (e.g. facial expressions, facial attributes, or even colors).

GAN-based approaches for image translation have been actively studied. One of
the first proposals \cite{isola2017image}, which was capable of learning
consistent image domain transforms, employed a pair of images that could be
used to create models that convert from the original to the target domain (e.g.
segmentation labels to the original image). Unfortunately, this system requires
that both, images and target images must exist as pairs in the training dataset
in order to learn the transformation between domains. Several works
\cite{zhu2017unpaired,choi2018stargan} have tried to address this drawback.
They suggested to use the virtual result in the target domain. In this
way, if the virtual result is inverted again, the inverted result must match
with the original image.  In these works, the framework can flexibly control
the image translation into different target domains.

\section{Method}
In the following section, we describe ATI-GAN approach, which addresses
image-to-image translation for facial attribute transfer. We explain the
training of the reconstructor, generator and discriminator in detail, showing that our
model trains in an introspective manner, such that it can estimate the
difference between the generated (fake) samples and the real samples, and finally
update itself to produce more realistic samples. 

\subsection{Model Architecture}
The network architecture of our proposal is depicted in Fig. \ref{fig:all}. It
is separated in an inpainting network (reconstructor Fig. \ref{fig:r}), generative
network (generator Fig.\ref{fig:g}) and discriminative network (discriminator Fig.
\ref{fig:d}). By combining each of these blocks in a sequential manner, the
ensemble model is able to perform an end-to-end attribute transfer.\\

\begin{figure}[t]
\begin{center}   
\includegraphics[width=\linewidth]{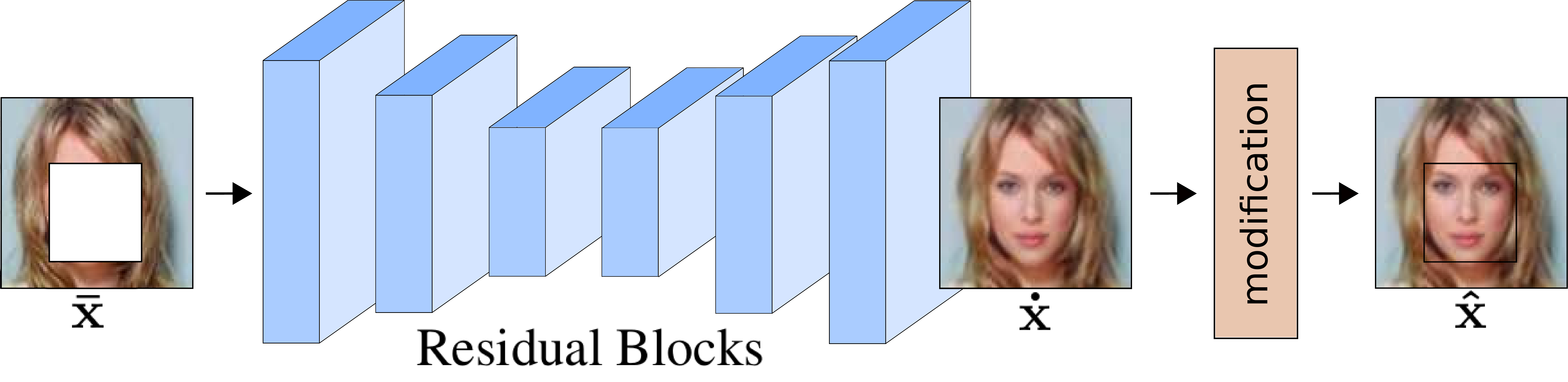}
\end{center}
   \caption{The figure shows the reconstructor structure with $\mathrm{\mathbf{\bar{x}}}$ as input and $\mathrm{\mathbf{\dot{x}}}$ as output. Also depicts the modification on $\mathrm{\mathbf{\dot{x}}}$ to get  $\mathrm{\mathbf{\hat{x}}}$, which will be fed into $D$.}
\label{fig:inp0}
\end{figure}

\noindent \textbf{Reconstructor.}
Given an image $\mathrm{\mathbf{x}}$ and its masked paired $\mathrm{\mathbf{\bar{x}}}$, the reconstructor $R$ takes $\mathrm{\mathbf{\bar{x}}}$ and tries to fill the large missing region with plausible content so that the output $\mathrm{\mathbf{\dot{x}}}$ looks realistic (see Fig. \ref{fig:inp0}). To achieve this objective, the reconstructor $R$ is backed up by the discriminator $D$ (as if it were a vanilla GAN setting) that assesses the reconstructed images. We can formulate the inpainting training process as a minimization problem 
\begin{align}
\begin{split}
	\mathcal{L}_{\mathrm{rec}} = \,& \lambda_{\mathrm{ae}} \mathcal{L}_{\mathrm{ae}}  \,+\, \mathcal{L}_{\mathrm{adv}} \,+\, \lambda_{\mathrm{c}}\,{\mathcal{L}_{\mathrm{class}}^f}.
\end{split}
\end{align}

\vspace{3mm}On the one hand, $\mathcal{L}_{\mathrm{ae}}$ constrains the reconstructed image by minimizing the absolute differences between the estimated values and the existing target values. It is defined as
\begin{align}
\begin{split}
	\mathcal{L}_{\mathrm{ae}} = \,& ||\mathrm{\mathbf{\bar{x}}} \,-\, \mathrm{\mathbf{\dot{x}_{ct}}}||_{1}  \,+\, \lambda_{\mathrm{p}} \, ||\mathrm{\mathbf{x_{p}}} \,-\, \mathrm{\mathbf{\dot{x}_{p}}}||_{1}
\end{split}
\end{align}

\begin{figure}[!t]
\centering
\begin{subfigure}[b]{0.47\textwidth}
   \hspace*{0.8cm} \includegraphics[width=.8\linewidth]{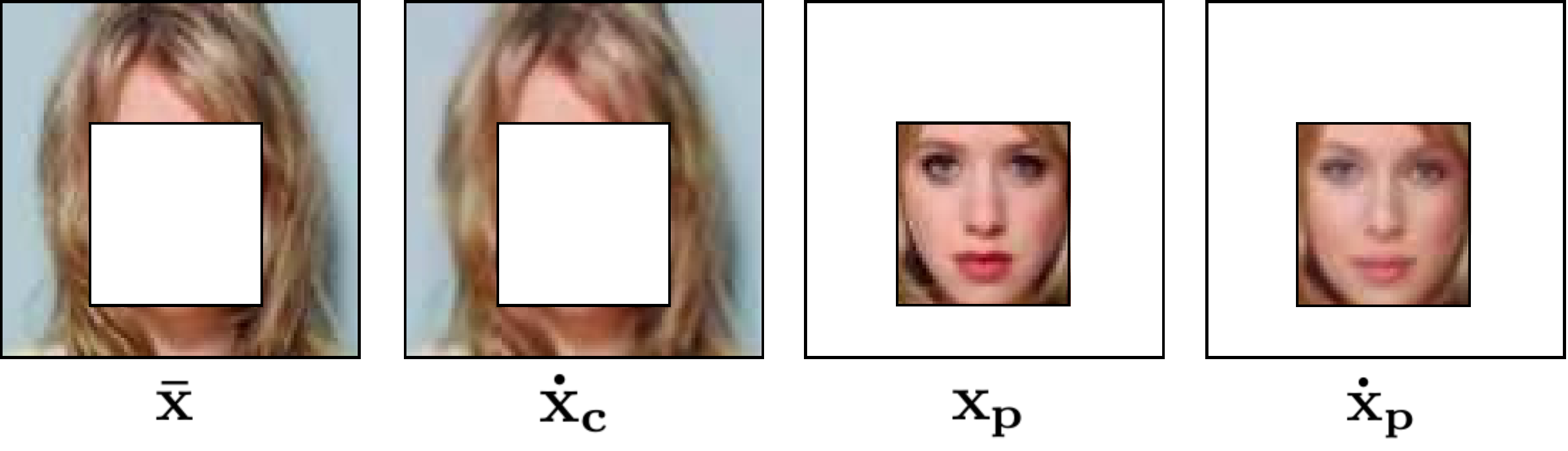}
   \caption{Elements from $\mathcal{L}_{\mathrm{ae}}$: masked image $\mathrm{\mathbf{\bar{x}}}$, contour from reconstructed image $\mathrm{\mathbf{\dot{x}_{ct}}}$, patch from original image $\mathrm{\mathbf{x_{p}}}$ and from reconstructed image $\mathrm{\mathbf{\dot{x}_{p}}}$.}
	\label{fig:patch}
\end{subfigure}

\begin{subfigure}[b]{0.47\textwidth}

   \vspace*{4mm} \hspace*{1.4cm} \includegraphics[width=.62\linewidth]{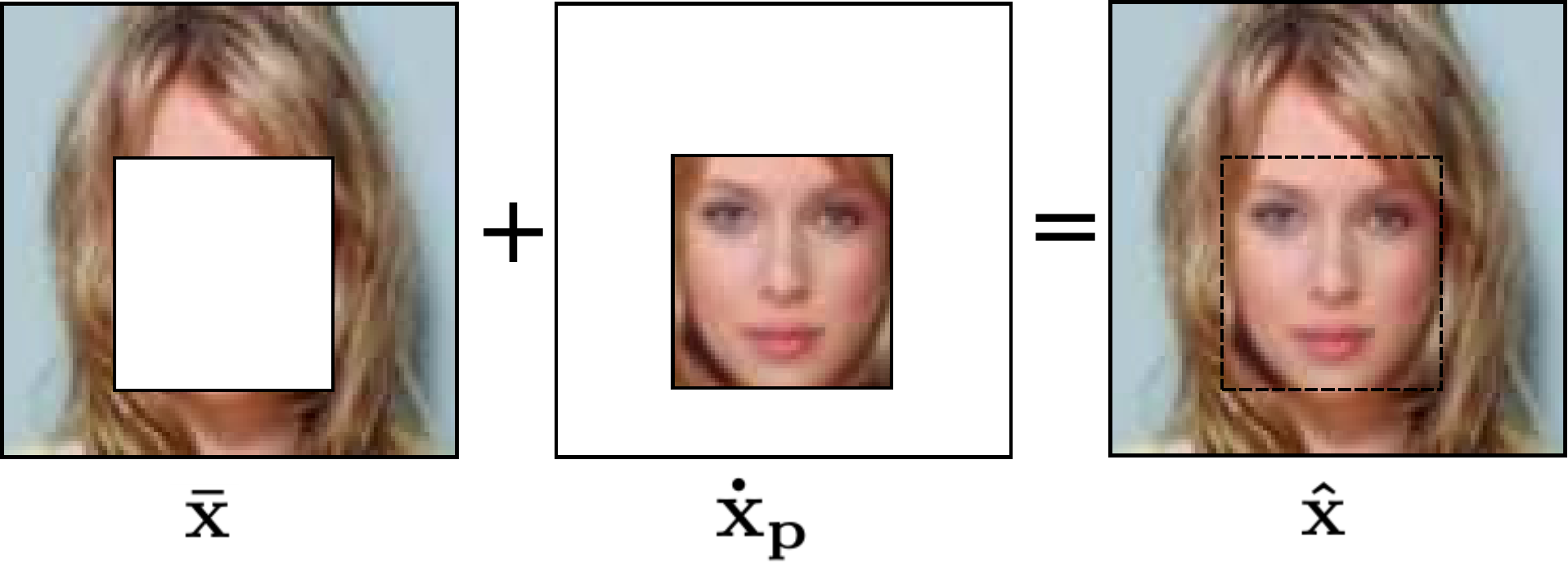}
   \caption{\textit{Modification} step: $\mathrm{\mathbf{\hat{x}}}$ is produced by combining $\mathrm{\mathbf{\bar{x}}}$ and $\mathrm{\mathbf{\dot{x}_{p}}}$.}
   	\label{fig:combi}
\end{subfigure}
\caption{Elements involved in the reconstructor training step.}
\end{figure}

\vspace{3mm}where the subindeces ct and p refer to contour and patch respectively. In Fig. \ref{fig:patch} is shown a visual example of the elements from $\mathcal{L}_{\mathrm{ae}}$ and in Fig. \ref{fig:inp0} the \textit{modification} step.

We apply on $\mathcal{L}_{\mathrm{ae}}$ a separately $l_1$  distance norm for the contour and for the patch. We treat them differently because the patch loss does not have to be strictly 0 since those synthetic images, which are not exactly equal to the original, are also a valid solution, if only if they look realistic and fit with the contour. 

On the other hand, $\mathcal{L}_{\mathrm{adv}}$ penalizes unrealistic images and ${\mathcal{L}_{\mathrm{class}}^f}$ (see Eq. \ref{eq:f}) looks after incorrect image-domain transformations. For these two cases the ideal solution converges to 0 loss. Further details are discussed in the following subsections.

In terms of topology implementation, we adopt the coarse network architecture introduced in \cite{yu2018generative}. Since the size of the receptive fields are a decisive factor in inpainting tasks, we use dilated convolutions to guarantee a sufficient large size. Additionally, we use mirror padding for all convolution layers and exponential linear unit (ELUs) as activation functions.\\

\begin{figure}[H]
\begin{center}
   \includegraphics[width=\linewidth]{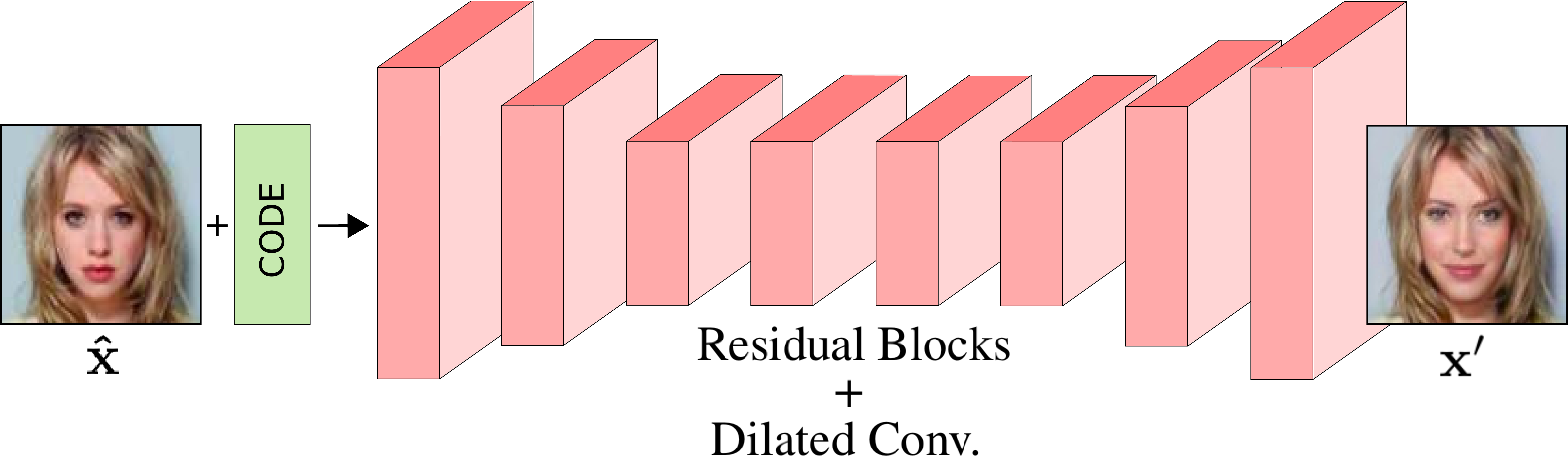}
\end{center}
   \caption{The figure shows the generator structure with $\mathrm{\mathbf{\hat{x}}}$ and feature code as inputs and $\mathrm{\mathbf{x'}}$ as output.}
\label{fig:gen0}
\end{figure}

\noindent \textbf{Generator.}
The role of the generator $G$ is to learn the mappings among multiple attribute
domains. To achieve this goal, we use the modified output from the
reconstructor $\mathrm{\mathbf{\hat{x}}}$  (see Fig. \ref{fig:combi}) as input.
Then we train $G$ to translate $\mathrm{\mathbf{\hat{x}}}$ into an output image $\mathrm{\mathbf{x'}}$,
 which is conditioned on the target domain code label $c_{\mathrm{target}}$ (see
Fig. \ref{fig:gen0}). In the same manner as in $R$, both
$\mathcal{L}_{\mathrm{adv}}$ and ${\mathcal{L}_{\mathrm{class}}^f}$ also play
important roles in the generator loss definition. Moreover, we have an extra term called cycle consistency loss $\mathcal{L}_{cycle}$. Previous works
\cite{kim2017learning,zhu2017unpaired,choi2018stargan} have shown how this term
helps to create strong paths between latent space and outcomes, and in our
case, it guarantees that the translated images $\mathrm{\mathbf{x'}}$ preserve the
content of their input images $\mathrm{\mathbf{\hat{x}}}$  while changing only
the domain-related part of the inputs though the latent space. This term is
defined as
\begin{align}
\begin{split}
	 \mathcal{L}_{cycle} = \,& ||\mathrm{\mathbf{\hat{x}}} \,-\, \mathrm{\mathbf{x''}} ||_{1}
\end{split}
\end{align}

\vspace{3mm}where 
\begin{align}
\begin{split}
	\mathrm{\mathbf{x'}} = \,& G(\mathrm{\mathbf{\hat{x}}}, c_{\mathrm{target}}) \\ \mathrm{\mathbf{x''}} = \,& G(\mathrm{\mathbf{x'}}, c_{\mathrm{original}}).
\end{split}
\end{align}

\vspace{3mm}Finally, joining it all terms together we have the following formula for generator loss
\begin{align}
\begin{split}
	\mathcal{L}_{\mathrm{gen}} =  \,&  \mathcal{L}_{\mathrm{adv}}  \,+\, \lambda_{\mathrm{c}}\,{\mathcal{L}_{\mathrm{class}}^f}\,+\, \lambda_{\mathrm{cycle}} \, \mathcal{L}_{cycle}.
\end{split}
\end{align}

\vspace{3mm}Note that the generator performs the entire cyclic translation $\mathrm{\mathbf{\hat{x}}}\rightarrow \mathrm{\mathbf{x'}} \rightarrow \mathrm{\mathbf{x''}}$ for every sample, forcing the code to be crucial for moving among domains. First, the original image $\mathrm{\mathbf{\hat{x}}}$ is translated into $\mathrm{\mathbf{x'}}$ (target domain), and then $\mathrm{\mathbf{x'}}$ is translated back to the original domain as $\mathrm{\mathbf{x''}}$.

Regarding the topology, we construct our baseline generator based on one of state-of-the-art image-to-image translation model \cite{choi2018stargan}, which is an adaptation from \cite{zhu2017unpaired}. On top of it, we apply several topology changes to adapt it to our system. In the aftermath, the final generator architecture consists of three convolutional layers, two of them with the stride size of two for down-sampling, six residual blocks, where in each block dilated layers with different dilatation
values are added, and two transposed convolutional layers with the stride size
of two for upsampling. We use instance normalization in the network.\\

\noindent \textbf{Discriminator.}
Our discriminator $D$ behaves slightly different from vanilla GAN. It takes samples of real and generated data and then tries to classify them correctly according to their attribute. Additionally, there is a second in-built classifier that tries to determine the domain from each sample.  As a result, the discriminator needs to have one adversarial loss that judges the appearance of the images $\mathcal{L}_{\mathrm{adv}}$ and one classification loss that classifies the attributes $\mathcal{L}_{\mathrm{class}}$.

\begin{figure}[!t]
\begin{center}
   \includegraphics[width=\linewidth]{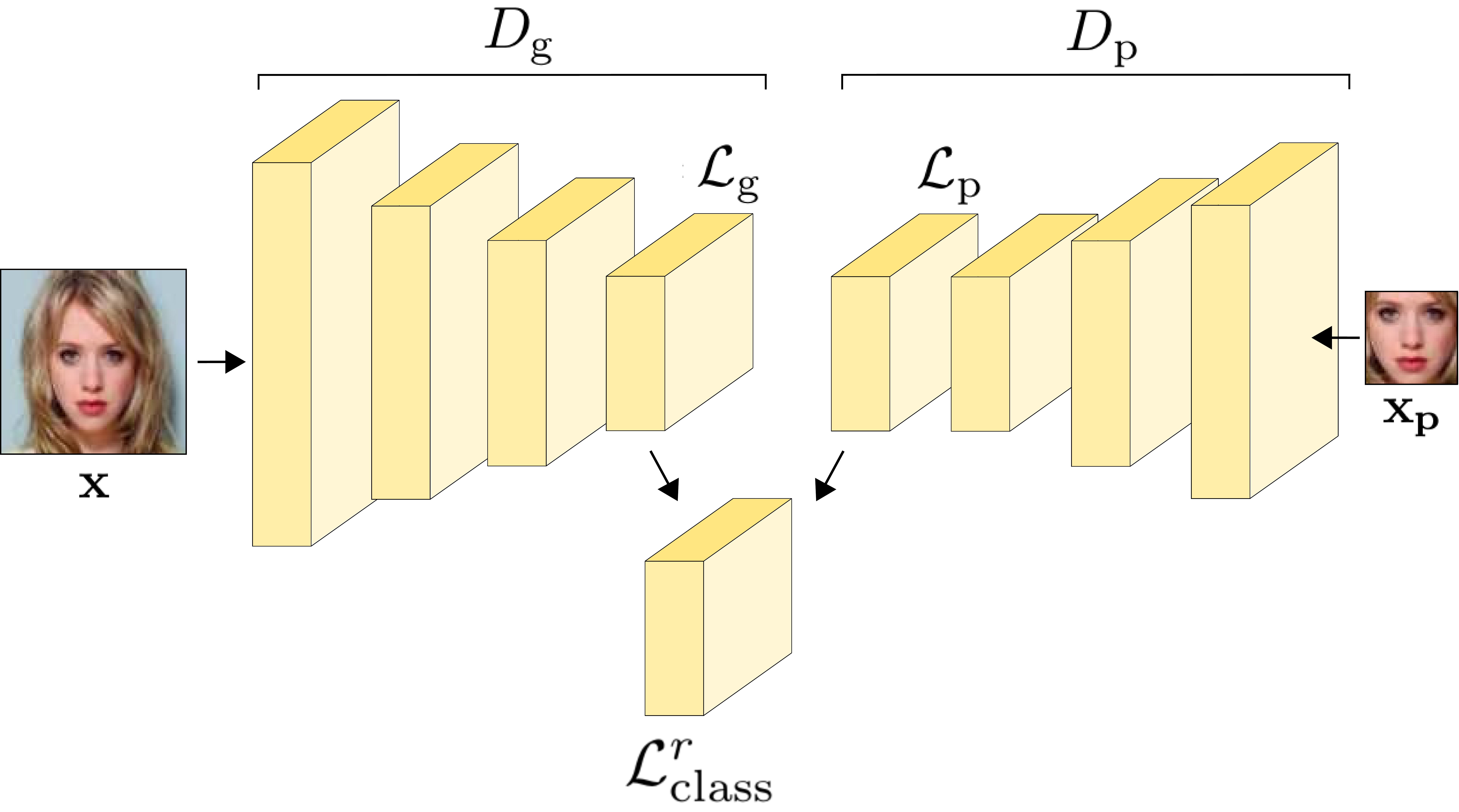}
\end{center}
   \caption{Illustration of the inner structure of our discriminator. It has two parts: $D_{\mathrm{g}}$ which discriminates at global scale (producing $\mathcal{L}_{\mathrm{g}}$) and $D_{\mathrm{p}}$ which does so at patch scale (producing $\mathcal{L}_{\mathrm{p}}$). There is also the class error (either $\mathcal{L}_{\mathrm{class}}^r$ or $\mathcal{L}_{\mathrm{class}}^f$) which uses the whole $D$ architecture to determine the class label. Note that this figure shows an example of $D$ being trained and fed with real data and therefore, the superindex $r$. }
\label{fig:disc2}
\end{figure}

The inner structure of our discriminator $D$ does also differ from standard discriminators. It is split into two fully convolutional topologies $D_{\mathrm{g}}$ and $D_{\mathrm{p}}$, and one final convolutional layer to combine both discriminators' outputs (see  Fig. \ref{fig:disc2}). All layers from  $D$ have a stride size of two for downsampling followed by LeakyReLUs as activation function. 

While the global discriminator $D_{\mathrm{g}}$ assesses the semantic consistency of the whole image, the patch discriminator $D_{\mathrm{p}}$ deals with the reconstructed initial masked part to enforce local consistency. Consequently, every image is evaluated by two independent loss functions $\mathcal{L}_{\mathrm{p}}$ and $\mathcal{L}_{\mathrm{g}}$ which together form the adversarial loss
\begin{align}
\begin{split}
	\mathcal{L}_{\mathrm{adv}} = \,& \mathcal{L}_{\mathrm{g}} \,+\, \mathcal{L}_{\mathrm{p}}.
\end{split}
\end{align}

\vspace{3mm}Over the last year, formulations of adversarial loss functions have continuously been changing and improving. GANs based on Earth-Mover distance loss \cite{arjovsky2017wasserstein} were one of the first attempts to clearly outperform vanilla GAN \cite{goodfellow2014generative}. Consequently, several approaches on image-to-image translation \cite{mirza2014conditional,isola2017image,zhu2017unpaired} and generative inpainting networks \cite{iizuka2017globally,li2017generative,pathak2016context} relied on DCGAN \cite{radford2015unsupervised} for their adversarial supervision. However, more recent research \cite{gulrajani2017improved} has showed that there are beneficial effects for image generation when adding a gradient penalty instead of weight clipping to enforce the Lipschitz constraint. As a result, a second wave of publications \cite{choi2018stargan, yu2018generative} proposes to use WGAN-GP. Following this approach,  we write $\mathcal{L}_{\mathrm{g}}$ and $\mathcal{L}_{\mathrm{p}}$ as
\begin{align}
\begin{split}
	\mathcal{L}_{\mathrm{g}} = \,& \mathbb{E}_{\mathrm{\mathbf{x}}}[D_{g}(\mathrm{\mathbf{x}})] \,-\, \mathbb{E}_{\mathrm{\mathbf{\hat{x}}},c_{\mathrm{target}}}[D_{g}(G(\mathrm{\mathbf{\hat{x}}}, c_{\mathrm{target}}))] \\
	-\,&  \lambda_{\mathrm{gp}} \, \mathbb{E}_{\mathrm{\mathbf{\tilde{x}}}}[(||\nabla_{\mathrm{\mathbf{\tilde{x}}}} \,D_{g}(\mathrm{\mathbf{\tilde{x}}})||_{2}-1)^2],
\end{split}
\label{eq:wgan}
\end{align}

\vspace{3mm}where $\mathrm{\mathbf{\tilde{x}}}$ is sampled uniformly along a straight line between a pair of a real and generated images. $\mathcal{L}_{\mathrm{g}}$ is analog to $\mathcal{L}_{\mathrm{p}}$ after replacing $D_{\mathrm{g}}$ for $D_{\mathrm{p}}$.

As we mentioned above, similarly to \cite{chongxuan2017triple}, our discriminator $D$ relies on a second loss function $\mathcal{L}_{\mathrm{class}}$ which accounts for domain classification. It computes the binary cross entropy function between the output domain labels from $D$ and one of two variants, either $c_{\mathrm{target}}$ or $c_{\mathrm{original}}$. Therefore, we can write the classification loss training from the reconstructor or the generator as
\begin{align}
\begin{split}
	{\mathcal{L}_{\mathrm{class}}^f} = \,& \mathbb{E}_{\mathrm{\mathbf{\hat{x}}},c_{\mathrm{target}}}[-\log D(c_{\mathrm{target}}|G(\mathrm{\mathbf{\hat{x}}}, c_{\mathrm{target}}) )],
\end{split}
\label{eq:f}
\end{align}

\vspace{3mm}where $f$ stands for fake input. Note that if we break Eq. \ref{eq:f} down, we can see that the input from $D$ is the result of $G$ when given a reconstructed input image $\mathrm{\mathbf{\hat{x}}}$ and a target domain label $c_{\mathrm{target}}$. In the case of training the reconstructor, $D$ will be directly fed with $\mathrm{\mathbf{\hat{x}}}$ but this time conditioned on $c_{\mathrm{original}}$. 
On the other hand, the discriminator loss is written as
\begin{align}
\begin{split}
	{\mathcal{L}_{\mathrm{class}}^r} = \,& \mathbb{E}_{\mathrm{\mathbf{\hat{x}}},c_{\mathrm{original}}}[-\log D(c_{\mathrm{original}}|\mathrm{\mathbf{x}})],
\end{split}
\end{align}

\vspace{3mm}where $r$ stands for real input. Note that the procedure is almost the same as for the reconstructor, but now the input in $D$ is the real image $\mathrm{\mathbf{x}}$, and by extension, we call it ${\mathcal{L}_{\mathrm{class}}^r}$.

The key of a good global training relies heavily on the discriminator. On the one hand, $D$ indirectly forces the generator by penalizing the gradients through ${\mathcal{L}_{\mathrm{class}}^f}$ to produce correct image-domain transformations as well as it learns to generate the correct output label when is fed with real data ${\mathcal{L}_{\mathrm{class}}^r}$. On the other hand, it also needs to learn to classify between the true and generated data  and penalize accordingly $\mathcal{L}_{\mathrm{adv}}$. Hence, the discriminator's optimization problem is formulated as
\begin{align}
\begin{split}
	\mathcal{L}_{\mathrm{disc}} =  \,&  -\mathcal{L}_{\mathrm{adv}} \,+\, \lambda_{\mathrm{c}}\,{\mathcal{L}_{\mathrm{class}}^r}.
\end{split}
\end{align}

%

\section{Experiments}
In this section, we present results for a series of experiments evaluating the proposed method, both quantitatively and qualitatively. We first give a detailed introduction of the experimental setup. Then, we discuss the inpainting outcomes and finally, we review the attribute transfer results.

\subsection{Experimental Settings}
We train ATI-GAN on the CelebFaces Attributes (CelebA) dataset \cite{liu2015deep}. It consists of 202,599 celebrity face images with variations in facial attributes. We randomly select 2,000 images for testing and use all remaining images as the training dataset. In training, we crop and resize the initially 178x218 pixel image to 128x128 pixels, and we mask them with 52x52 size patches. These masked regions are always centered around the tip of the nose, occluding in most of the cases a large portion of the face. All experiments presented in this paper have been conducted on a single NVIDIA GeForce GTX 1080 GPU, without applying any post-processing.

\subsection{Training}
Since our model is divided into three distinguishable parts (reconstruction/inpainting, generative and discriminative), three independent Adam optimizer with $\beta_{1} = 0.5$, $\beta_{2} = 0.999$. are used during training. We set the batch size to 16 and run the experiments for 200,000 iterations. We start using the output of the reconstructor as input for the generator after iteration 50K. In this way, we ensure that the gradient updates in the generative model are reliable. We update the generator after every five discriminator updates as in \cite{gulrajani2017improved, choi2018stargan}. The learning rate used in the implementation is 0.0001 for the first 10 epochs and  then linearly decreased to 0 over the next 10 epochs. The losses are weighted by the factors: $\lambda_{\mathrm{ae}}$, $\lambda_{\mathrm{cycle}}$ and  $\lambda_{\mathrm{gp}}$ set to 10,  $\lambda_{\mathrm{p}}$ to 5 and $\lambda_{\mathrm{c}}$ to 1 respectively. The training procedure as a whole is described in Algorithm \ref{alg:main}.

\begin{algorithm}
 \caption{Training of the proposed architecture. In all experiments were used the following default values  $n_{\mathrm{iter}} = 200,000$, $th_{\mathrm{disc}}= \frac{n_{\mathrm{iter}}}{4}$, $\alpha_{\mathrm{disc}} = \alpha_{\mathrm{rec}} = \alpha_{\mathrm{gen}} = 0.0001$, $m = 16$, $n_{\mathrm{gen}}=5.$ }
 \label{alg:main}
 \begin{algorithmic}[1]
  \STATE Require 1: $n_{\mathrm{iter}}$, number of iterations. $th_{\mathrm{disc}}$, threshold indicating when $G$ starts to use modified output from reconstructor ($\mathrm{\mathbf{\hat{x}}}$). $\alpha$'s, learning rates. $m$, batch size. $n_{\mathrm{gen}}$, number of iterations of the generator per discriminator iteration.

  \STATE Require 2: $w_{0}$, initial discriminator parameters. $\theta_{0}$, initial reconstructor parameters. $\gamma_{0}$, initial generator parameters.
  \FOR {$i < n_{\mathrm{iter}}$}
  
  \STATE Sample $\{\mathrm{\mathbf{x}}^{(j)}\}_{j=0}^m$ a batch from real data
  \STATE Sample $\{\mathrm{\mathbf{\bar{x}}}^{(j)}\}_{j=0}^m$ a batch from masked data
  
  \STATE $\#$ Train discriminator $D$
  \STATE $w_{\mathrm{disc}} \leftarrow w_{\mathrm{disc}} - \alpha_{\mathrm{disc}} \nabla_{w} \{ \mathcal{L}_{\mathrm{disc}}(\mathrm{\mathbf{x}},G(\mathrm{\mathbf{\hat{x}}})) \}$
  
  \STATE $\#$ Train reconstructor $R$
  \STATE $\theta_{\mathrm{rec}} \leftarrow \theta_{\mathrm{rec}} - \alpha_{\mathrm{rec}} \nabla_{\theta} \{ \mathcal{L}_{\mathrm{rec}}(\mathrm{\mathbf{x}},\mathrm{\mathbf{\bar{x}}})\}$
  \STATE $\#$ Train generator $G$
  \IF {$mod(i,n_{\mathrm{gen}}) = 0$}
  \IF {$ i < th_{\mathrm{disc}}$}
  \STATE $\gamma_{\mathrm{gen}} \leftarrow \gamma_{\mathrm{gen}} - \alpha_{\mathrm{gen}} \nabla_{\gamma} \{\mathcal{L}_{\mathrm{gen}}(\mathrm{\mathbf{x}})\}$
  \ELSE
  \STATE $\gamma_{\mathrm{gen}} \leftarrow \gamma_{\mathrm{gen}} - \alpha_{\mathrm{gen}} \nabla_{\gamma} \{\mathcal{L}_{\mathrm{gen}}(\mathrm{\mathbf{\hat{x}}})\}$
  \ENDIF
  \ENDIF
  
  \ENDFOR
 \end{algorithmic} 
 \end{algorithm}

\subsection{Image Inpainting}
The image inpainting problem has a number of different formulations. The definition of our interest is: given that most of the pixels of a face are unobserved because they are masked, our objective is to restore them in a natural way, so that we end up having a plausible and realistic face. In order to achieve good inpaiting results, our synthesized faces must fit into these masks/holes taking into account both, the reconstruction quality of the face as well as the adaptation with the rest of the image. Note that this pixel transformation will be conditioned on the desired attribute transfer too.

As mentioned in \cite{yeh2017semantic,yu2018generative}, it is important to notice that there is no perfect numerical metric for semantic inpainting due to the existence of infinite amount of possible solutions. Note that image inpainting algorithms do not try to reconstruct the ground-truth image, but to fill the masked area with content that looks realistic.  As a result, the ground-truth image is only one of many possibilities. Following classical inpainting approaches, we employ in our evaluation study the peak signal-to-noise ratio (PSNR). However, this metric might oversimplify the comparison since it directly measures difference in pixel values. Therefore, it is usually combined with a second metric called  structural similarity (SSIM) which offers a more elaborated and reliable measurement values. 

 \begin{table}[!h]
\caption{Quantitative evaluation in terms of PSNR and SSIM metrics on the CelebA testing dataset. Higher values are better.}
\centering
\resizebox{0.73\linewidth}{!}{\begin{tabular}{ccc}
\hline
 Method & PSNR (dB) & SSIM \\
\hline
 SIIWGAN \cite{vitoria2018semantic} & 19.20 & 0.920 \\
 SIIDGM \cite{yeh2017semantic} & 19.40 & 0.907 \\
 CE \cite{pathak2016context} & 21.30 & 0.923 \\
 GL \cite{iizuka2017globally} & 23.19 & 0.936 \\
 GntInp\cite{yu2018generative} & 23.80 & 0.940 \\
 GMCNN \cite{wang2018image} & 24.46 & 0.944 \\
 GL+LID\cite{li2019generative} & 25.56 & \textbf{0.953} \\
 ours & \textbf{31.80} & 0.946 \\
\hline
\end{tabular}}
\label{table:psnr}
\end{table}

Our inpainting goal is always under the same conditions i.e. regenerating missing facial attributes, our mask will be a central square patch in the image. This is the standard crop procedure for CelebA since most of the information lies on the center of the image. Table \ref{table:psnr} shows the comparison results for PSNR and SSIM metrics, where similar works have also reported the scores based on square centered crops.

We are inclined to think that the improvement of the metrics (specially PSNR) comes from a good equilibrium between our reconstructor and our discriminator. While the $R$ learns to produce the coarse features from faces (natural-looking structures) via $\mathcal{L}_{\mathrm{ae}}$, $D$ enforces to smooth the results ``asking'' for finer details via $\mathcal{L}_{\mathrm{adv}}$ through $D_{\mathrm{g}}$ and $D_{\mathrm{p}}$.

Finally, theses results demonstrate that our approach is able to utilize the end-to-end model architecture to propagate informative gradients, which eventually lead to a significant performance gain. Nevertheless, note that we do not aim at outperforming  state-of-the-art image inpainting techniques, but we use it as a crucial part for our attribute transfer system.

%

\subsection{Attribute Transfer}
In this subsection, we focus on attribute manipulation. We validate that faces change according to the specified target attribute. This phenomenon is known as attribute transfer or morphing.  In particular, we focus on the following set of attributes: \textit{eyeglasses}, \textit{mustache}, \textit{smiling} and \textit{young}.

\begin{figure}[!t]
\begin{subfigure}[t]{\linewidth}
\centering
   \includegraphics[width=1\linewidth]{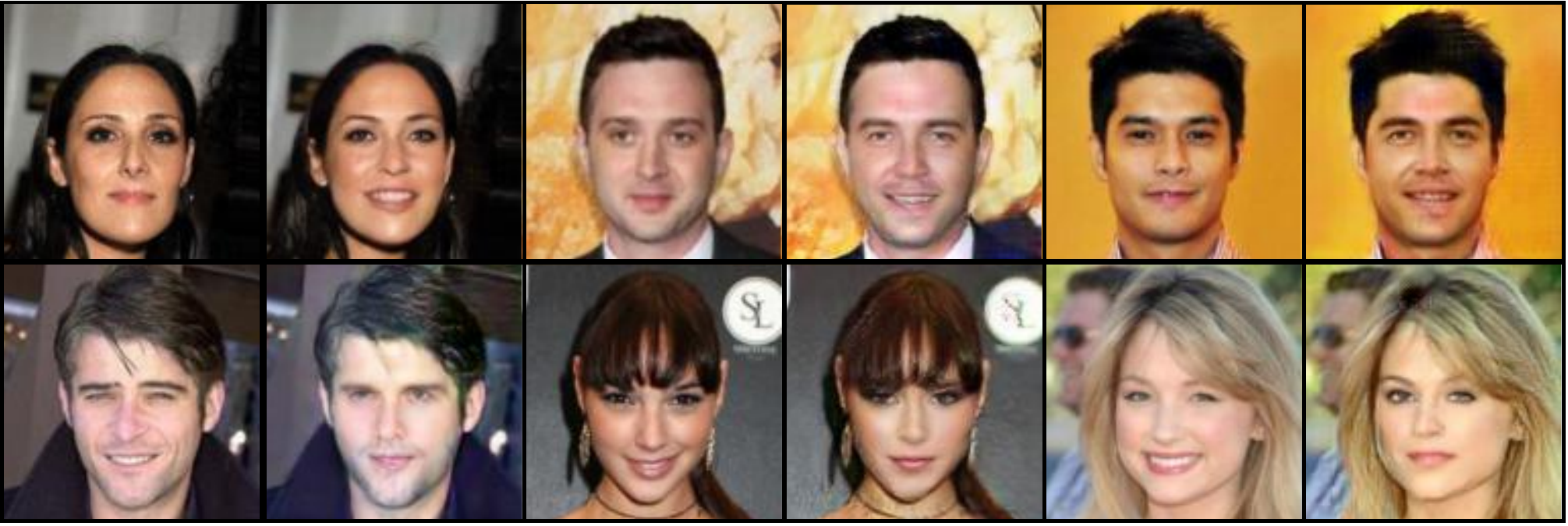}
   \caption{Smiling (up) and Not-Smiling (down) transformations.}
   \label{fig:exp21} 
\end{subfigure}\\[1.25ex]
\begin{subfigure}[t]{\linewidth}
\centering
	\includegraphics[width=1\linewidth]{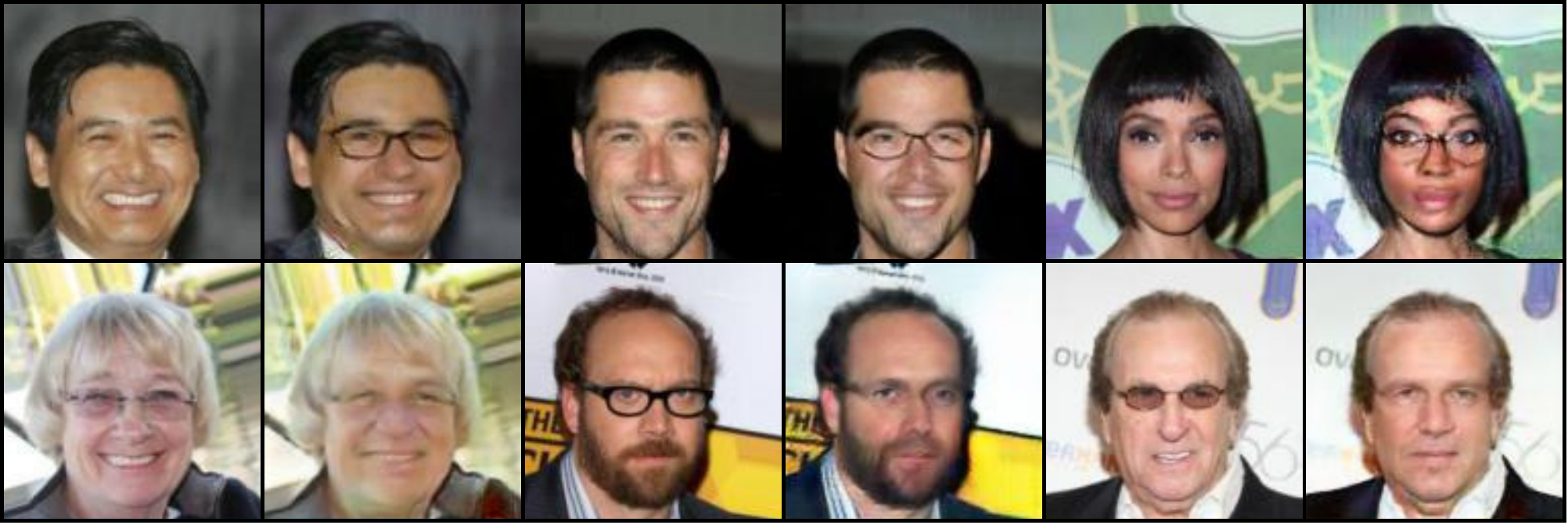}
	\caption{Eyeglasses (up) and Not-Eyeglasses (down) transformations.}
	\label{fig:exp22}
\end{subfigure}\\[1.25ex]
\begin{subfigure}[t]{\linewidth}
\centering
   \includegraphics[width=1\linewidth]{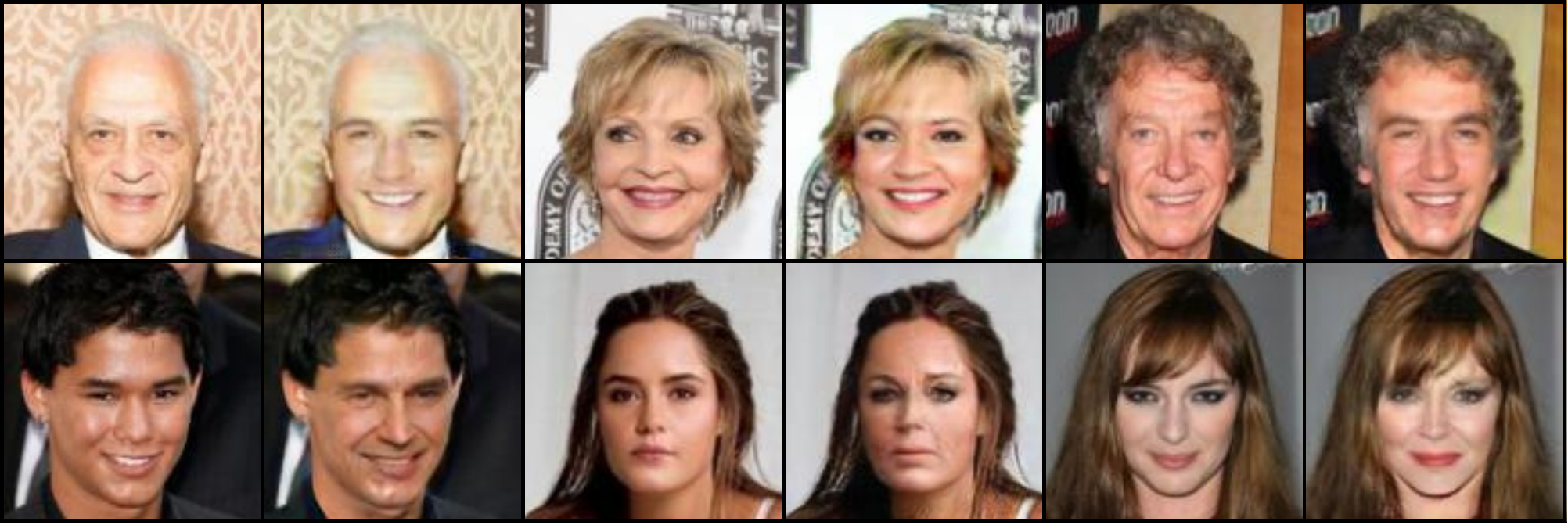}
   \caption{Young (up) and Old (down) transformations.}
   \label{fig:exp23}
\end{subfigure}\\[1.25ex]
\begin{subfigure}{\linewidth}
\centering
   \includegraphics[width=1\linewidth]{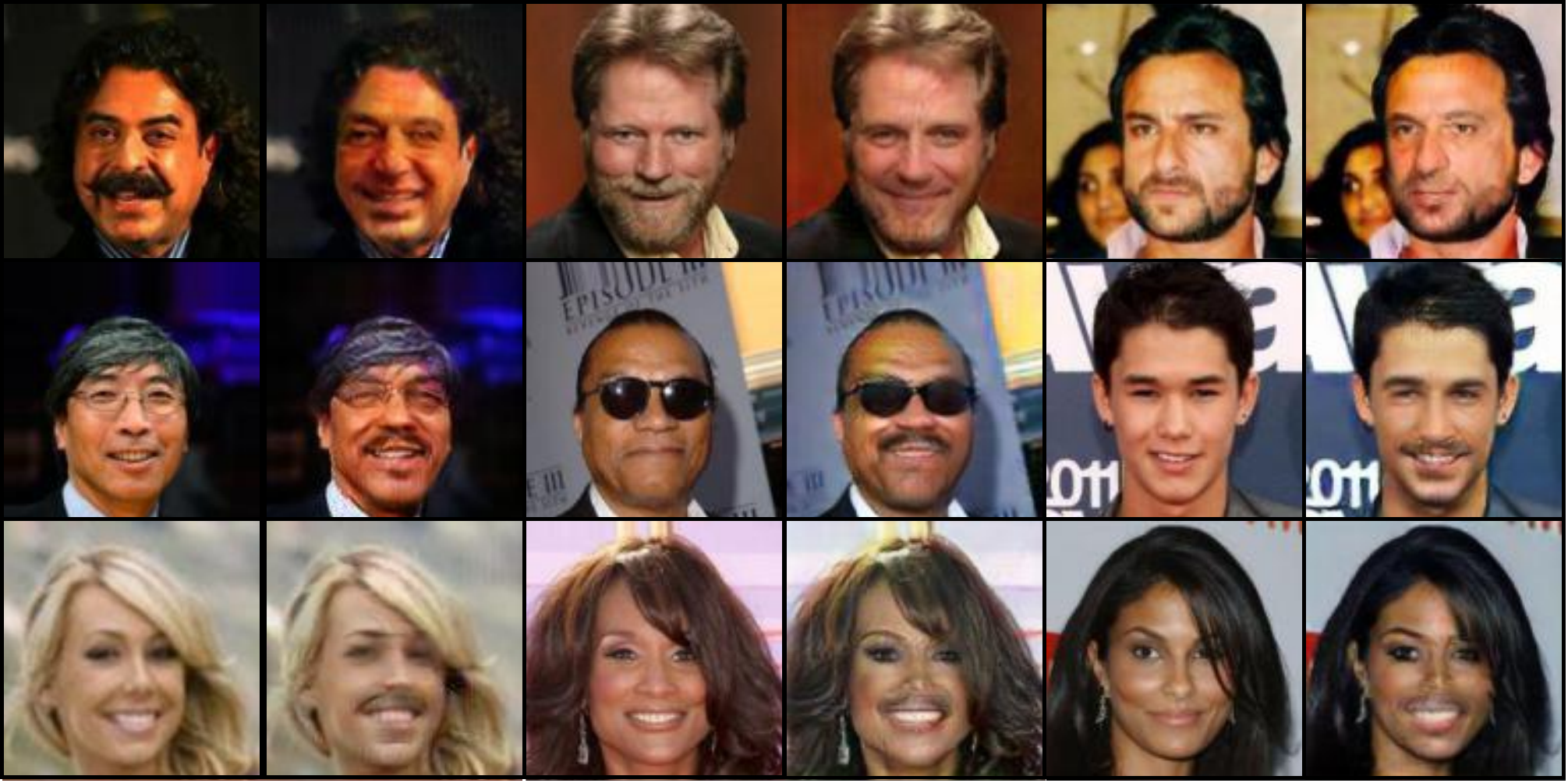}
   \caption{Not-Mustache (up) and Mustache (down) transformations.}
   \label{fig:exp25}
\end{subfigure}
\caption{Each pair depicts the real image (left) and the transformed image (right). The variety within the examples shows robust feature translation independently of genre, race and age.}
\label{fig:exp2}
\end{figure}

Figure \ref{fig:exp2} shows the transformed images (with the target attribute), in which we can qualitatively determine the results of the model by judging the attribute transfer results. We can observe that ATI-GAN clearly generates natural-looking faces containing the target attributes providing very competitive results on test data. This is
possible because of the inherent properties of the end-to-end system that takes advantage of the inpainting structure (among others) presented in this work.

Additionally, we have performed a user study to assess attribute transfer tasks. It consists of a survey in which users have to label with 0 when the attribute is not recognized and 1 otherwise. For each attribute transfer, we conduct the test on $10\%$ of the testing data. Note that the generated images have a single attribute translation from the aforementioned list.

\begin{table}[!h]
\caption{Perceptual evaluation on transformed images for each attribute (\textit{smiling}, \textit{eyeglasses}, \textit{young} and \textit{mustache}).}
\centering
\resizebox{\linewidth}{!}{\begin{tabular}{|c|c|}
\hline
Smiling $\rightarrow$ Not-Smiling & Not-Smiling $\rightarrow$ Smiling\\
\hline
86$\%$ &  88$\%$ \\
\hline
\hline
Eyeglasses $\rightarrow$ Not-Eyeglasses & Not-Eyeglasses $\rightarrow$ Eyeglasses\\
\hline
50$\%$ &  82$\%$\\
\hline
\hline
Young $\rightarrow$ Old & Old $\rightarrow$ Young\\
\hline
85$\%$ &  70$\%$\\
\hline
\end{tabular}}\\[0.8ex]

\centering
\resizebox{\linewidth}{!}{\begin{tabular}{|c|c|c|c|}
\hline
\multicolumn{2}{|c|}{Mustache $\rightarrow$ Not-Mustache} & \multicolumn{2}{c|}{Not-Mustache $\rightarrow$ Mustache}\\
\hline
\multicolumn{2}{|c|}{Men} & \hspace*{0.55cm}Men\hspace*{0.55cm} & Women\\
\hline
\multicolumn{2}{|c|}{$33\%$} & \hspace*{0.55cm}$98\%$\hspace*{0.55cm} & $68\%$\\
\hline 
\end{tabular}}
\label{table:transfer}
\end{table}

\begin{figure}[H]
\begin{center}
   \includegraphics[width=\linewidth]{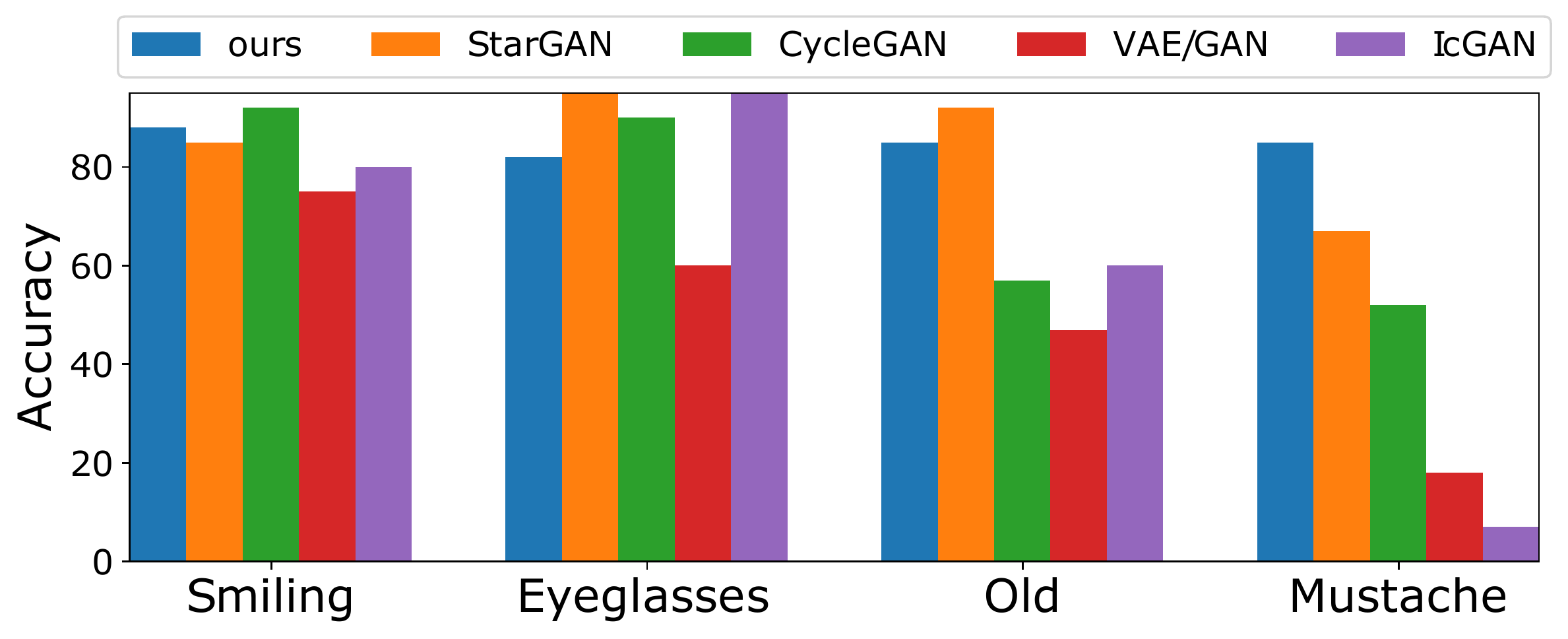}
\end{center}
   \caption{Comparisons among StarGAN \cite{choi2018stargan}, CycleGAN \cite{zhu2017unpaired},VAE/GAN \cite{larsen2016autoencoding}, IcGAN \cite{perarnau2016invertible} and an our approach.}
\label{fig:comp}
\end{figure}

According to Table \ref{table:transfer}, a big part of our translations achieves a successful attribute transformation. More interestingly, however, is to analyze what the meaning behind the percentage is. For instance, we observe that the transformations Eyeglasses $\rightarrow$  Not-Eyeglasses and vice versa have no symmetry on transferring attribute. This mainly happens because ATI-GAN is not an invertible model, therefore, moving from domain A to B involves one path and from B to A another path. Additionally, since CelebA suffers from unbalanced attributes, meaning that not all the attributes are equally present, the ability to transfer might be conditioned on the amount of samples containing the involved feature. For example, Not-Mustache $\rightarrow$ Mustache has much lower success rate for women than for men, because there are no examples of women wearing mustaches. Finally, Fig. \ref{fig:comp} shows the the attribute editing
accuracy of different models.

\section{Ablation Study}

In this section, we quantitatively evaluate the impact of the reconstructor block $R$. Additionally, we compare our end-to-end approach to other attribute transfer methods.

\begin{table}[!h]
\caption{Quantitative results on CelebA.}
\centering
\resizebox{0.73\linewidth}{!}{\begin{tabular}{ccc}
\hline
 Method & PSNR (dB) & SSIM \\
\hline
 StarGAN \cite{choi2018stargan} & 22.80 & 0.819 \\
 AttGAN \cite{he2019attgan} & 24.07 & 0.841 \\
 GraphCNN \cite{bhattarai2020inducing} & 27.20 & 0.897 \\
 \hline
 ours (w/o $R$) & 29.84 & 0.901 \\
 ours & \textbf{31.80} & \textbf{0.946} \\
\hline
\end{tabular}}
\label{table:ablation}
\end{table}

\section{Conclusions}
In this paper, we introduce a novel image-to-image translation model capable of
applying an accurate local attribute transformation. Previous attribute transfer works were mostly based only on the manipulation of GAN latent space. However, we propose a completely different approach utilizing inpainting as a part of our embedded system. Our method takes advantage of the fact that attributes are induced by local structures. Therefore, it is sufficient to change only parts of the image, while the remaining parts can be used to force the generator into realistic outputs. We show how ATI-GAN can synthesize high quality human face images. We do believe the method is generalisable to other objects and domains being able to produce synthetic data containing certain
attributes on demand. We see many interesting avenues of future work including exploring multi-attribute transfer.

{\small
\bibliographystyle{IEEEtran}
\bibliography{bare_conf}
}

\end{document}